\documentclass[10pt,twocolumn,letterpaper]{article}

\usepackage[final]{cvpr}
\usepackage{graphicx}
\usepackage{amsmath}
\usepackage{amssymb}
\usepackage{booktabs}
\usepackage{makecell}
\usepackage{multirow}
\usepackage{multicol}
\usepackage{comment}
\usepackage[accsupp]{axessibility}
\usepackage[pagebackref,breaklinks,colorlinks]{hyperref}

\usepackage[capitalize]{cleveref}
\crefname{section}{Sec.}{Secs.}
\Crefname{section}{Section}{Sections}
\Crefname{table}{Table}{Tables}
\crefname{table}{Tab.}{Tabs.}

\def\cvprPaperID{7932}
\def\confName{CVPR}
\def\confYear{2022}

\begin{document}

\title{\emph{SimMIM}: a Simple Framework for Masked Image Modeling}

\author{
Zhenda Xie$^1$\thanks{Equal. Zhenda, Yutong, Zhuliang are long-term interns at MSRA.}
\quad Zheng Zhang$^2$\textsuperscript{*} \quad Yue~Cao$^2$\textsuperscript{*} \\
\quad Yutong Lin$^3$ \quad Jianmin Bao$^2$
\quad Zhuliang Yao$^1$ \quad Qi Dai$^2$
\quad Han~Hu$^2$\textsuperscript{*} \\
{$^1$Tsinghua University} \quad  {$^2$Microsoft Research Asia} \quad {$^3$Xi'an Jiaotong University} \\
\small{\texttt{\{t-zhxie,zhez,yuecao,t-yutonglin,jianmin.bao,t-zhuyao,qid,hanhu\}@microsoft.com}}
}

\maketitle

\begin{abstract}

 This paper presents SimMIM, a simple framework for masked image modeling. We have simplified recently proposed relevant approaches, without the need for special designs, such as block-wise masking and tokenization via discrete VAE or clustering. To investigate what makes a masked image modeling task learn good representations, we systematically study the major components in our framework, and find that the simple designs of each component have revealed very strong representation learning performance: 1) random masking of the input image with a moderately large masked patch size (e.g., 32) makes a powerful pre-text task; 2) predicting RGB values of raw pixels by direct regression performs no worse than the patch classification approaches with complex designs; 3) the prediction head can be as light as a linear layer, with no worse performance than heavier ones. Using ViT-B, our approach achieves 83.8\% top-1 fine-tuning accuracy on ImageNet-1K by pre-training also on this dataset, surpassing previous best approach by +0.6\%. When applied to a larger model with about 650 million parameters, SwinV2-H, it achieves 87.1\% top-1 accuracy on ImageNet-1K using only ImageNet-1K data. We also leverage this approach to address the data-hungry issue faced by large-scale model training, that a 3B model (SwinV2-G) is successfully trained to achieve state-of-the-art accuracy on four representative vision benchmarks using $40\times$ less labelled data than that in previous practice (JFT-3B). The code is available at \url{https://github.com/microsoft/SimMIM}.

\end{abstract}

\section{Introduction}
\label{sec:intro}

``\emph{What I cannot create, I do not understand.}''

~~~~~~~~~~~~~~~~~~~~~~~~~~~~~~~~~~~~~~~~~~~~~~~~~~~~~---~Richard Feynman

``Masked signal modeling'' is one such task that learns to create: masking a portion of input signals and trying to predict these masked signals. In NLP, following this philosophy, self-supervised learning approaches built on the masked language modeling tasks have largely repainted the field~\cite{devlin2018bert,liu2019roberta,brown2020language}, i.e., learning very large-scale language models by using huge amounts of unlabeled data has been shown to generalize well to a broad range of NLP applications.

\begin{figure}
    \centering
    \includegraphics[width=\linewidth]{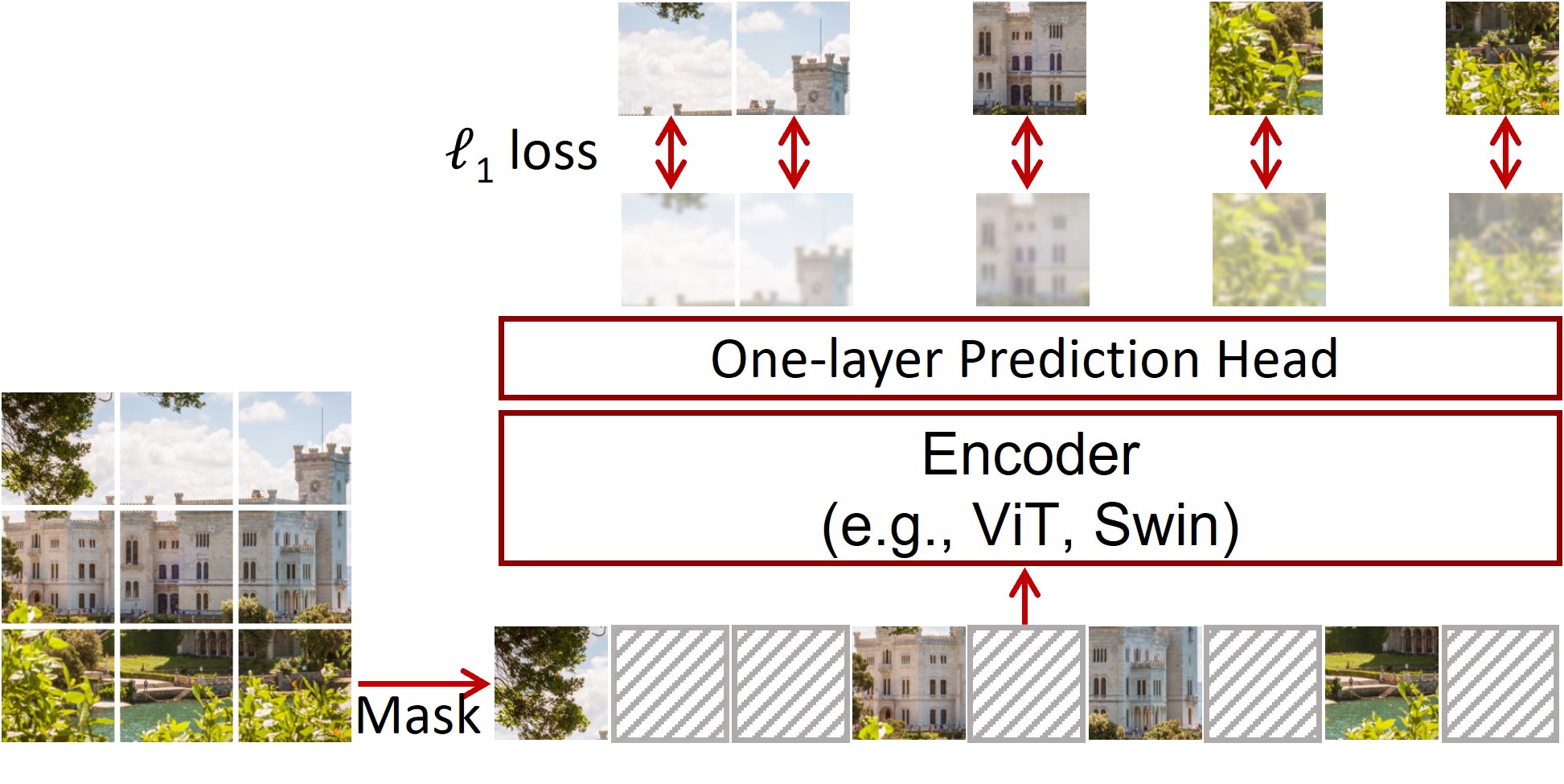}
    \vspace{-20pt}
    \caption{An illustration of our simple framework for masked language modeling, named \emph{SimMIM}. It predicts raw pixel values of the randomly masked patches by a lightweight one-layer head, and performs learning using a simple $\ell_1$ loss. }
    \label{fig:teaser}
    \vspace{-1em}
\end{figure}

In computer vision, although there are pioneers leveraging this philosophy for self-supervised representation learning~\cite{zhang2016colorization,zhang2017splitbrain,doersch2015context}, in previous years, this line of work was almost buried by the contrastive learning approaches~\cite{wu2018memorybank,he2019moco,chen2020simclr}. 
The different difficulties of applying this task to the language and visual domains can be explained by the differences between two modalities.
One of the differences is that images exhibit stronger locality: pixels that are close to each other tend to be highly correlated~\cite{hubel1962receptive}, so the task can be done by duplicating close pixels rather than by semantic reasoning.
Another difference is that visual signals are \emph{raw and low-level}, while text tokens are human-generated \emph{high-level} concepts. This raises a question of whether the prediction of low-level signals is useful for high-level visual recognition tasks.
A third difference is that the visual signal is \emph{continuous}, and the text token is \emph{discrete}. It is unknown how classification-based masked language modeling approaches can be adapted to handle continuous visual signals well.

Until recently, there have been trials that attempt to bridge modality gaps and resolve the obstacles, by introducing several special designs, for example, by converting continuous signals into color clusters~\cite{chen2020imagegpt}, by patch tokenization using an additional network~\cite{bao2021beit}, or by a block-wise masking strategy to break short-range connections ~\cite{bao2021beit}, etc. Through these special designs, the learned representations proved to be well transferable to several visual recognition tasks.

In contrast to requiring special complex designs, in this paper, we present a simple framework which aligns well with the nature of visual signals, as shown in Figure~\ref{fig:teaser}, and is able to learn similar or even better representations than previously more complex approaches: \emph{random masking of input image patches, using a linear layer to regress the raw pixel values of the masked area with an $\ell_1$ loss}. The key designs and insights behind this simple framework include:
\begin{itemize}
\vspace{-5pt}
    \item Random masking is applied on image patches, which is simple and convenient for vision Transformers. For masked pixels, either larger patch size or higher masking ratio can result in a smaller chance of finding visible pixels that are close. For a large masking patch size of 32, the approach can achieve competitive performance in a wide range of masking ratios (10\%-70\%). For a small mask patch size of 8, the masking ratio needs to be as high as 80\% to perform well. Note that the preferred masking ratios are very different from that in the language domain, where a small masking ratio of 0.15 is adopted as default. We hypothesize that different degrees of information redundancy in two modalities may lead to the different behaviors.
\vspace{-5pt}
    \item A raw pixel regression task is used. The regression task aligns well with the continuous nature of visual signals, which possesses ordering property. This simple task performs no worse than the classification approaches with classes specially defined by tokenization, clustering, or discretization.
\vspace{-5pt}
    \item An extremely lightweight prediction head (e.g., a linear layer) is adopted, which achieves similarly or slightly better transferring performance than that of heavier prediction heads (e.g., an inverse Swin-B). The use of an extremely lightweight prediction head brings a remarkable speedup in pre-training. In addition, we note that a broad range of target resolutions (e.g., $12^2$-$96^2$) perform competitive with the highest $192^2$. While heavier heads or higher resolutions generally result in greater generation capability, this greater capability does not necessarily benefit down-stream fine-tuning tasks.
\end{itemize}
\vspace{-5pt}

Though simple, the proposed \emph{SimMIM} approach is very effective for representation learning. Using ViT-B, it achieves 83.8\% top-1 fine-tuning accuracy on ImageNet-1K, surpassing previous best approach (\cite{bao2021beit}) by +0.6\%. \emph{SimMIM} has also shown to be scalable to larger models: with a SwinV2-H model (658M parameters)~\cite{swinv2}, it achieves 87.1\% top-1 accuracy on ImageNet-1K classification, which is the highest number among methods that use ImageNet-1K data only. This result encourages the use of self-supervised learning to address the increasing data-hungry problem caused by quickly rising model capacity. In fact, with the help of \emph{SimMIM}, we successfully trained a SwinV2-G model with 3 billion parameters~\cite{swinv2} using $\sim$40$\times$ smaller data than that of Google's JFT-3B dataset, and set new records on several representative benchmarks: 84.0\% top-1 accuracy on  ImageNet-V2 classification~\cite{recht2019imagenetv2}, 63.1/54.4 box/mask mAP on COCO object detection~\cite{chen2019htc,lin2014coco}, 59.9 mIoU on ADE20K semantic segmentation~\cite{zhou2018ade,xiao2018upernet}, and 86.8\% top-1 accuracy on Kinetics-400 action recognition~\cite{kay2017kinetics,liu2021video}.

While in recent years we have witnessed an increasing overlap between NLP and computer vision in both basic modeling and learning algorithms, as well as in multi-modal applications, which aligns well with how human brains achieve general intelligence capabilities, we hope that our demonstration of ``masked signal modeling'' in computer vision can drive this trend a bit further and encourage deeper interaction of different AI fields.

\section{Related Work}
\label{sec:related-work}

\paragraph{Masked language modeling (MLM)} Masked language modeling~\cite{devlin2018bert,liu2019roberta} and its auto-regressive variants~\cite{brown2020language} are the dominant self-supervised learning approaches in the field of natural language processing (NLP). Given visible tokens in a sentence or a sentence pair / triplet, the approaches learn representations by predicting invisible tokens of the input. This line of approaches has repainted the field since about 3 years ago~\cite{devlin2018bert}, that it enables the learning of very large language models and generalizes well on broad language understanding and generation tasks by leveraging huge data.

\vspace{-9pt}

\paragraph{Masked image modeling (MIM)} Masked image modeling~\cite{doersch2015context,pathak2016context,henaff2019CPC,trinh2019selfie,chen2020imagegpt} progressed in parallel with the MLM task in NLP but located in a non-mainstream position for a long time. 
The context encoder approach~\cite{pathak2016context} is a pioneer work in this direction, which masks a rectangle area of the original images, and predicts the missing pixels. 
CPC~\cite{henaff2019CPC,trinh2019selfie} predicts patches via a verification task in each batch with a contrastive predictive coding loss.
Recently, iGPT~\cite{chen2020imagegpt}, ViT~\cite{dosovitskiy2020vit} and BEiT~\cite{bao2021beit} recall this learning approach on the modern vision Transformers, and show strong potential in representation learning by introducing special designs on some components, such as clustering on pixels~\cite{chen2020imagegpt}, prediction of mean color~\cite{dosovitskiy2020vit}, and tokenization via an additional dVAE network with a block-wise masking strategy~\cite{bao2021beit}. 
In contrary to these complex designs, we present an extremely simple framework, SimMIM, which shows similar or even slightly better effectiveness.

\vspace{-9pt}
\paragraph{Reconstruction based methods} are also related to our approach, particularly the auto-encoder approaches~\cite{HinSal06,NIPS2006_87f4d79e,vincent2008dae,vincent2010stackeddae,kingma2013vae,oord2017neural}. Similar as in our approach, they adopt a reconstruction task to recover the original signals. However, they are based on a different philosophy of visible signal \emph{reconstruction}, other than the \emph{creation or prediction} of invisible signals as in our approach. They thus progress in a very different path, by studying how to effectively regularize the task learning by proper regularization or architecture bottlenecks.

\vspace{-9pt}
\paragraph{Image inpainting methods} Beyond representation learning, masked image modeling is a classical computer vision problem, named image inpainting. This problem has been extensively studied in computer vision for a long time~\cite{perez2003poisson,Yeh_2017_CVPR,yu2019free}, aiming for improving the inpainting quality and without connecting to self-supervised representation learning. While we advocate image inpainting as a strong self-supervised pre-text task, we also find stronger inpainting capability does not necessarily leads to stronger fine-tuning performance on down-stream tasks.

\vspace{-9pt}
\paragraph{Compressed sensing} The approach in this paper is also related to compressed sensing~\cite{donoho2006compressed}, which affirms most of the data we acquire including image signals can be thrown away with almost no perceptual loss. Such claim is also partly supported by recent works of sparse inference~\cite{han2021dynamic} that the recognition accuracy has very little drop after throwing a large portion of image features ~\cite{huang2017multi,xie2020spatially,riquelme2021moe}. 
The observation in this paper goes further for the input signals, that with an extremely small portion of randomly selected input image patches as input, i.e., 10\%, the inpainting task can still be learnt to produce good visual representations. 

\vspace{-9pt}
\paragraph{Other self-supervised learning approaches}
During the last two decades, there have been numerous pretext tasks to learn visual representation in a self-supervised way: gray-scale image colorization~\cite{zhang2016colorization}, jigsaw puzzle solving~\cite{noroozi2016jigsaw}, split-brain auto-encoding~\cite{zhang2017splitbrain}, rotation prediction~\cite{gidaris2018rotation}, learning to cluster~ \cite{caron2018deepcluster}. Though very different from masked image modeling, some of them interestingly also follow a philosophy of predicting the invisible parts of signals, e.g., \cite{zhang2016colorization,zhang2017splitbrain} use one or two color channels as input to predict values of other channels. Another large portion of works lie in the contrastive learning approaches~\cite{dosovitskiy2014exemplarcnn,wu2018memorybank,he2019moco,chen2020simclr,cao2020pic,grill2020byol,xie2021pixpro}, which are the previous mainstream. We hope our work can encourage the study of masked language modeling as a pre-text task for self-supervised visual representation learning.

\section{Approach}
\label{sec:approach}
\subsection{A Masked Image Modeling Framework}

Our approach \emph{SimMIM} learns representation through masked image modeling, which masks a portion of input image signals and predicts the original signals at masked area. The framework consists of 4 major components:
\begin{itemize}
\vspace{-5pt}
    \item[1)] \emph{Masking strategy}. Given an input image, this component designs how to select the area to mask, and how to implement masking of selected area. The transformed image after masking will be used as the input.
\vspace{-5pt}
    \item[2)] \emph{Encoder architecture}. It extracts a latent feature representation for the masked image, which is then used to predict the original signals at the masked area. The learnt encoder is expected to be transferable to various vision tasks. In this paper, we mainly consider two typical vision Transformer architectures: a vanilla ViT~\cite{dosovitskiy2020vit} and Swin Transformer~\cite{liu2021swin}.
\vspace{-5pt}
    \item[3)] \emph{Prediction head}. The prediction head will be applied on the latent feature representation to produce one form of the original signals at the masked area. 
\vspace{-5pt}
    \item[4)] \emph{Prediction target}. This component defines the form of original signals to predict. It can be either the raw pixel values or a transformation of the raw pixels. This component also defines the loss type, with typical options including the cross-entropy classification loss and the $\ell_1$ or $\ell_2$ regression losses.
\end{itemize}
\vspace{-5pt}

In the following subsections, we will present typical options of each component. These options are then systematically studied. By combining simple designs of each component, we have been able to achieve strong representation learning performance.

\subsection{Masking Strategy}

\begin{figure*}
    \centering
    \includegraphics[width=\linewidth]{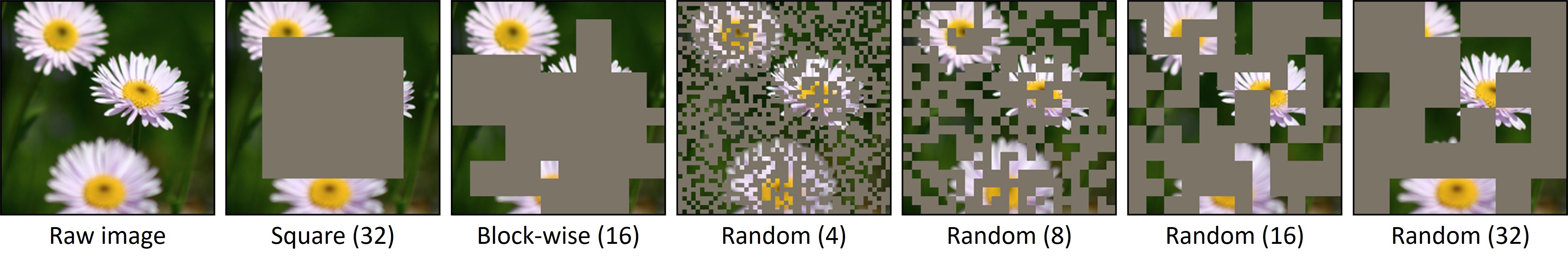}
    \vspace{-23pt}
    \caption{Illustration of masking area generated by different masking strategies using a same mask ratio of 0.6: square masking~\cite{pathak2016context}, block-wise masking~\cite{bao2021beit} apply on 16-sized patches, and our simple random masking strategy on different patch sizes (e.g., 4, 8, 16 and 32).}
    \label{fig:mask_type}
    \vspace{-10pt}
\end{figure*}

For input transformation of masked area, we follow the NLP community~\cite{devlin2018bert,liu2019roberta} and BEiT~\cite{bao2021beit} to use a learnable mask token vector to replace each masked patch. The token vector dimension is set the same as that of the other visible patch representation after patch embedding. For masking area selection, we study the following masking strategies (illustrated in Figure~\ref{fig:mask_type}):

\vspace{-5pt}
\paragraph{Patch-aligned random masking} We first present a patch-aligned random masking strategy. Image patches are the basic processing units of vision Transformers, and it is convenient to operate the masking on patch-level that a patch is either fully visible or fully masked. For Swin Transformer, we consider equivalent patch sizes of different resolution stages, 4$\times$4$\sim$32$\times$32, and adopt 32$\times$32 by default which is the patch size of the last stage. 
For ViT, we adopt 32$\times$32 as the default masked patch size.

\vspace{-5pt}
\paragraph{Other masking strategies} We also try other masking strategies in previous works: 1)  \cite{pathak2016context} introduces a central region masking strategy. We relax it to be randomly movable on the image. 2) \cite{bao2021beit} introduces a complex block-wise masking strategy. We try this mask strategy on two masked patch sizes of $16\times 16$ and $32\times 32$.

\subsection{Prediction Head}

The prediction head can be of arbitrary form and capacity, as long as its input conforms with the encoder output and its output accomplishes the prediction target. Some early works follow auto-encoders to employ a heavy prediction head (decoder)~\cite{pathak2016context}. 
In this paper, we show that the prediction head can be made extremely lightweight, as light as \emph{a linear layer}. We also try heavier heads such as a 2-layer MLP, an inverse Swin-T, and an inverse Swin-B. 

\subsection{Prediction Targets}

\paragraph{Raw pixel value regression} The pixel values are continuous in the color space. A straight-forward option is to predict raw pixels of the masked area by regression. In general, vision architectures usually produce feature maps of down-sampled resolution, e.g., $16\times$ in ViT and $32\times$ for most other architectures. To predict all pixel values at a full resolution of input images, we map each feature vector in feature map back to the original resolution, and let this vector take charge of the prediction of corresponding raw pixels.

For example, on the $32\times$ down-sampled feature maps produced by a Swin Transformer encoder, we apply a $1\times 1$ convolution (linear) layer with output dimension of $3072 = 32\times 32\times 3$ to stand for the RGB values of $32\times 32$ pixels. We also consider lower resolution targets by downsampling the original images by $\{32\times, 16\times, 8\times, 4\times, 2\times\}$, respectively.

An $\ell_1$-loss is employed on the masked pixels:
\begin{equation}
    L = \frac{1}{\Omega(\mathbf{x}_M)}\|\mathbf{y}_M - \mathbf{x}_M\|_1,
\end{equation}
where $\mathbf{x},\mathbf{y} \in \mathbb{R}^{3HW\times 1}$ are the input RGB values and the predicted values, respectively; $M$ denotes the set of masked pixels; $\Omega(\cdot)$ is the number of elements. We also consider $\ell_2$ and smooth-$\ell_1$ loss in experiments which perform similarly well, and $\ell_1$ loss is adopted by default.

\vspace{-5pt}
\paragraph{Other prediction targets}
Previous approaches mostly convert the masked signals to clusters or classes, and then perform a classification task for masked image prediction. 
\begin{itemize}
\vspace{-5pt}
    \item \emph{Color clustering}. In iGPT~\cite{chen2020imagegpt}, the RGB values are grouped into 512 clusters by $k$-means using a large amount of natural images. Each pixel is then assigned to the closest cluster center. This approach requires an additional clustering step to generate the 9-bit color palette. In our experiments, we use the 512 cluster centers learnt in iGPT.
\vspace{-5pt}
    \item \emph{Vision tokenization}. In BEiT~\cite{bao2021beit}, a discrete VAE (dVAE) network~\cite{ramesh2021zero} is employed to transform image patches to dVAE tokens. The token identity is used as the classification target. In this approach, an additional dVAE network needs to be pre-trained.
\vspace{-5pt}
    \item \emph{Channel-wise bin color discretization}. The R, G, B channels are separately classified, with each channel discretized into equal bins, e.g., 8 and 256 bins used in the experiments.
\end{itemize}
\vspace{-5pt}

\subsection{Evaluation protocols}
We follow~\cite{bao2021beit} to mainly evaluate the quality of learnt representations by fine-tuning on ImageNet-1K image classification, which is a more usable scenario in practice. We will mainly account for this metric in our ablations. 
In the system-level comparison, we also follow previous works~\cite{he2019moco,chen2020simclr,cao2020pic,grill2020byol,chen2020imagegpt,bao2021beit} to report the performance on previous dominant metric of linear probing. Nevertheless, we will not account on this linear probing metric, as our main goal is to learn representations which can well complement the following down-stream tasks.

\section{Experiments}
\label{sec:exp}

\subsection{Ablation Study}
\subsubsection{Settings}
We adopt Swin-B~\cite{liu2021swin} as the default backbone in our ablation study, which  facilitates us to evaluate the learnt representations also on downstream tasks such as object detection and semantic segmentation (see Appendix). To reduce experimental overhead, we use a default input image size of $192^2$ and adapt the window size as 6 to accommodate the changed input image size. The ImageNet-1K image classification dataset is used for both pre-training and fine-tuning.

In self-supervised pre-training, we employ an AdamW optimizer~\cite{kingma2014adam} with a \emph{cosine} learning rate scheduler, and train for 100 epochs. The training hyper-parameters are: the batch size as 2048, base learning rate as 8e-4, weight decay as 0.05, $\beta_1=0.9$, $\beta_2=0.999$, warm-up for 10 epochs. A light data augmentation strategy is used: random resize cropping with scale range of $[0.67, 1]$ and a aspect ratio range of $[3/4, 4/3]$, followed by a random flipping and a color normalization steps.

The default options for the components of SimMIM are: a random masking strategy with a patch size of 32$\times$32 and a mask ratio of 0.6; a linear prediction head with a target image size of $192^2$; an $\ell_1$ loss for masked pixel prediction. Our ablation is conducted by varying one option and keeping other settings the same as that of the default.

\begin{table}\small
    \centering
    \addtolength{\tabcolsep}{0.pt}
    \begin{tabular}{c|c|c|c}
    \toprule
    Mask & Masked & Mask  & Top-1 \\
    Type & patch size & ratio & acc (\%) \\
    \hline
    \multirow{3}{*}{\shortstack{square}} & 32 & 0.11 (2$\times$2) & 82.6 \\
     & 32 & 0.25 (3$\times$3) & 82.5 \\
     & 32 & 0.44 (4$\times$4) & 82.5 \\
    \hline
    \multirow{3}{*}{\shortstack{block-wise}} & 16/32 & 0.4 & 82.7/82.7\\
     & 16/32 & 0.6 & 82.6/82.6\\
     & 16/32 & 0.8 & 82.4/82.5\\
    \hline
    \multirow{5}{*}{\shortstack{random}} & 4/8/16/32 & 0.4 & 81.9/82.0/82.4/82.9 \\
     & 4/8/16/32 & 0.6 & 82.0/82.1/82.7/82.8 \\
     & 4/8/16/32 & 0.8 & 82.1/82.4/82.8/82.4 \\
    & 64 & 0.1 & 82.6 \\
     & 64 & 0.2 & 82.6 \\
    \hline
    \multirow{9}{*}{\shortstack{random}} & 32 & 0.1 & 82.7\\
     & 32 & 0.2 & 82.8\\
     & 32 & 0.3 & 82.8\\
     & 32 & 0.4 & 82.9\\
     & 32 & 0.5 & \textbf{83.0} \\
     & 32 & 0.6 & 82.8 \\
     & 32 & 0.7 & 82.7 \\
     & 32 & 0.8 & 82.4 \\
     & 32 & 0.9 & 82.4 \\
    \bottomrule
    \end{tabular}
    \vspace{-7pt}
    \caption{Ablation on different masking strategies (i.e., square, block-wise, and random) with different masked patch sizes (i.e., 4, 8, 16, 32 and 64).}
    \label{table:mask_strategy}
    \vspace{-10pt}
\end{table}

In fine-tuning, we also employ an AdamW optimizer, 100-epoch training, and a cosine learning rate scheduler with 10-epoch warm-up. The fine-tunig hyper-parameters are: the batch size as 2048, a base learning rate of 5e-3, a weight decay of 0.05, $\beta_{1}=0.9$, $\beta_{2}=0.999$, a stochastic depth~\cite{huang2016deep} ratio of 0.1, and a layer-wise learning rate decay of 0.9. We follow the same data augmentation used in \cite{bao2021beit}, including RandAug~\cite{cubuk2020randaugment}, Mixup~\cite{zhang2017mixup}, Cutmix~\cite{yun2019cutmix}, label smoothing~\cite{szegedy2016rethinking}, and random erasing~\cite{zhong2020random}.

\subsubsection{Masking Strategy}\label{sec:mask_strategy}
We first study how different masking strategies affect the effectiveness of representation learning. The fine-tuning accuracy of different approaches under multiple masking ratios are summarized in Table~\ref{table:mask_strategy}. 

\begin{figure}
    \centering
    \includegraphics[width=1.0\linewidth]{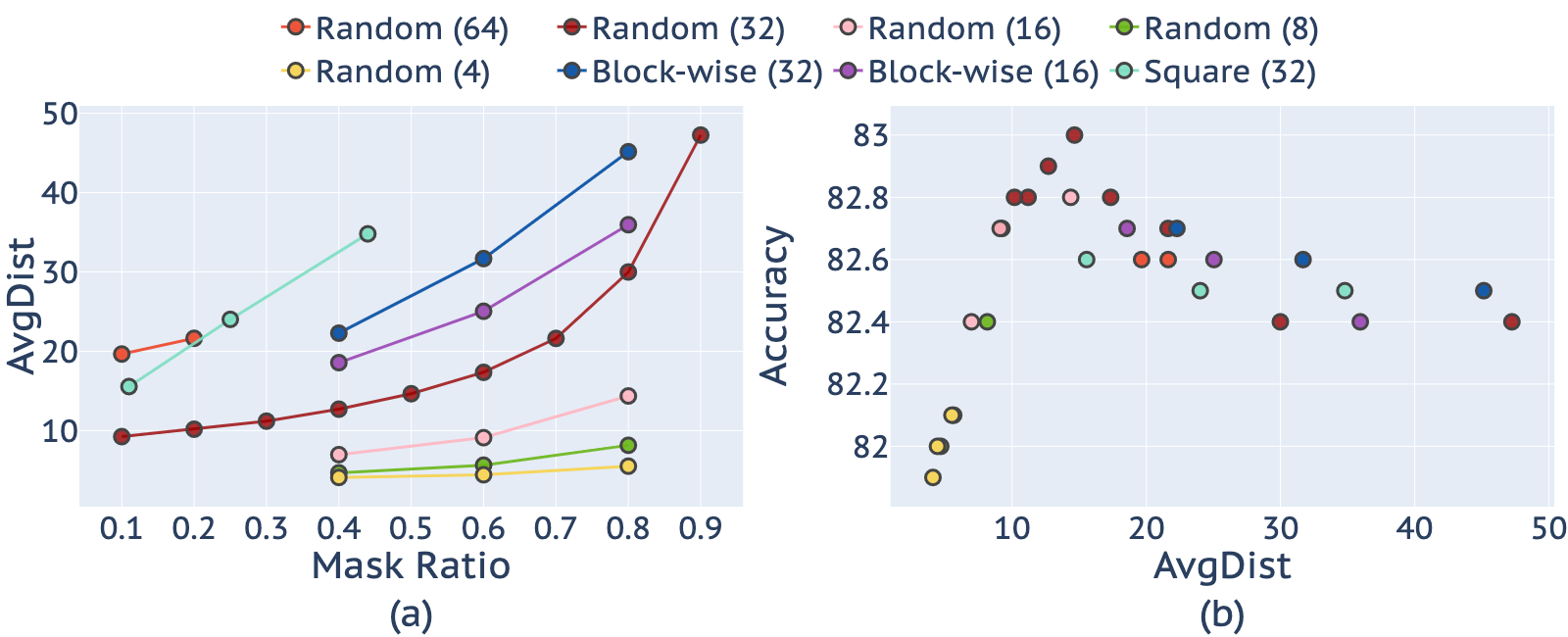}
    \vspace{-20pt}
    \caption{(a) \emph{AvgDist} (averaged distance of masked pixels to the nearest visible pixels) w.r.t. different masking ratios using different masking strategies and different masked patch sizes; (b) fine-tuning performance (top-1 accuracy) w.r.t. \emph{AvgDist}.}
    \label{fig:avg_min_dist}
    \vspace{-10pt}
\end{figure}

\begin{table}\small
    \centering
    \begin{tabular}{c|ccc}
    \toprule
    Head & \#params & Training costs & Top-1 acc (\%) \\ 
    \hline
    Linear & 89.9M & 1$\times$ & 82.8\\
    2-layer MLP & 90.9M & 1.2$\times$ & 82.8 \\
    inverse Swin-T & 115.2M & 1.7$\times$ & 82.4 \\
    inverse Swin-B & 174.8M & 2.3$\times$ & 82.5 \\
    \bottomrule
    \end{tabular}
    \vspace{-7pt}
    \caption{Ablation on different prediction heads. A simple linear layer performs the best with lower training costs.}
    \label{table:pred_head}
    \vspace{-10pt}
\end{table}

We first notice that the best accuracy of our simple random masking strategy reaches 83.0\%, which is +0.3\% higher than the best of other more specially designed strategies such as the block-wise masking as in~\cite{bao2021beit}. 

In addition, when a large masked patch size of 32 is adopted, this simple strategy performs stably well on a broad range of masking ratios (10\%-70\%). 
We hypothesize that the centering pixel of a large masked patch may be distant enough to visible pixels. Thus it enforces the network to learn relatively long-range connections, even when a low masking ratio is used (e.g., 10\%) or all patches around are not masked. 
Another way to increase the prediction distance is to use larger masking ratio, which also shows to benefit the fine-tuning performance of relatively small patch sizes. 
By increasing the masking ratio from 0.4 to 0.8 at a patch size of 4, 8 and 16, the accuracy is smoothly improved by +0.2\% (from 81.9\% to 82.1\%), +0.4\% (from 82.0\% to 82.4\%), and +0.4\% (from 82.4\% to 82.8\%), respectively. 
Nevertheless, the overall accuracy at these smaller patches is not as high as that at a larger patch size of 32. Further increasing the patch size to 64 is observed with degraded accuracy, probably due to the too large prediction distance.

The above observations and analyses can also be well reflected by a newly proposed \emph{AvgDist} metric, which measures the averaged Euclidean distance of masked pixels to the nearest visible ones. The \emph{AvgDist} of different masking strategies w.r.t. varying masking ratios are shown in Figure~\ref{fig:avg_min_dist}(a). From this figure, we observe that the \emph{AvgDist} of all masking strategies is smoothly increased with growing masking ratios. For random masking strategy, when the masked patch size is low, e.g., 4 or 8, the \emph{AvgDist} is relatively low and grows slowly with increasing masking ratios. On the other hand, when the patch size is large, e.g., 64, very small masking ratio (e.g. 10\%) still makes relatively large \emph{AvgDist}. The square and block-wise methods produce similarly high \emph{AvgDist} values as of patch size 64.

Figure~\ref{fig:avg_min_dist}(b) plots the relationship between fine-tuning accuracy and the \emph{AvgDist} measure, which follows a ridge shape. The entries of high fine-tuning accuracy roughly distribute in a range of $[10, 20]$ of \emph{AvgDist}, while entries with smaller or higher \emph{AvgDist} perform worse. This indicates that the prediction distance in masked image modeling is encouraged to be moderate, neither too large nor too small. Probably, small distance in masked prediction may let the network learn too much short connections, while large distance may be too difficult to learn. These results also indicate that \emph{AvgDist} may be a good indicator for the effectiveness of masked image modeling.

In our experiments, we adopt a masking ratio of 0.6 on patch size of 32 by default, due to its stable performance. Also note that the masking strategies and ratios in the language domain are very different from what explored in our work, which usually adopts a small masking ratio of 15\%. We hypothesize that different degrees of information redundancy by two modalities may lead to the different behaviors.

\subsubsection{Prediction Head}

Table~\ref{table:pred_head} ablates the effect of different prediction heads, including a linear layer, a 2-layer MLP, an inverse Swin-T and an inverse Swin-B. While generally heavier heads produce slightly lower losses, for example, 0.3722 (inverse Swin-B) versus 0.3743 (a linear layer), the transferring performances on the down-stream ImageNet-1K task are lower. It indicates that stronger inpainting capability does not necessarily result in better down-stream performance. It is probably because that the capacity is largely wasted in the prediction head, which will not be used in down-stream tasks. There is also a practical drawback, that a heavier prediction head brings higher training costs, e.g., the training cost of using an inverse Swin-B is 2.3$\times$ of that by a linear layer.

Also note that in previous contrastive learning approaches~\cite{he2019moco,chen2020simclr,grill2020byol}, it is a common practice to use a multi-layer MLP head in the pre-text tasks, instead of a linear layer, which makes the latent feature produced by the encoder moderately distant to the pre-text target, and shows beneficial for the linear probing evaluation metric. In our work, we show that a single linear layer head in our approach, under a fine-tuning metric, has shown competitive or even the optimal transferring performance. It indicates that if our aim is to learn good features for fine-tuning, the important exploration on head designing in contrastive learning approaches may not be necessary for that of masked image modeling.

\subsubsection{Prediction Resolution}

Table~\ref{table:pred_res} ablates the effect of varying target resolution. It shows that a large range of resolutions (e.g., $12^2$-$192^2$) perform equally well. The transferring performance drops only at a low resolution of $6^2$, probably because this option throws too much information away. These results imply the information granularity required by the down-stream image classification task. The effects to other more fine-grained down-stream tasks such as object detection or semantic segmentation will be explored in our future study.

Note that we adopt a default target resolution of $192^2$ in our experiments, due to the equally best transferring accuracy and the negligible computation overhead.

\begin{table}\small
    \centering
    \addtolength{\tabcolsep}{-1.pt}
    \begin{tabular}{c|c|c|c|c|c|c}
    \toprule
    \makecell[c]{Image size\\ \footnotesize{(ratio of inputs)}} & \makecell[c]{$6^2$\\ \footnotesize{(1/32)}} & \makecell[c]{$12^2$\\ \footnotesize{(1/16)}} & \makecell[c]{$24^2$\\ \footnotesize{(1/8)}} & \makecell[c]{$48^2$\\ \footnotesize{(1/4)}} & \makecell[c]{$96^2$\\ \footnotesize{(1/2)}} & \makecell[c]{$192^2$\\ \footnotesize{(1/1)}} \\ 
    \hline
    Top-1 acc (\%) & 82.3 & 82.7 & 82.8 & 82.7 & 82.8 & 82.8 \\
    \bottomrule
    \end{tabular}
    \vspace{-7pt}
    \caption{Ablation on different prediction resolutions. A moderately large resolution (no less than $1/16$ all perform well.}
    \label{table:pred_res}
    \vspace{-10pt}
\end{table}

\begin{table}\small
    \centering
    \begin{tabular}{c|c}
    \toprule
    Scope to predict & Top-1 acc (\%) \\
    \hline
    masked area & 82.8 \\
    full image & 81.7 \\
    \bottomrule
    \end{tabular}
    \vspace{-7pt}
    \caption{Ablation on different performing areas of prediction loss. If the loss is computed at masked area, it performs a pure prediction task. If it is computed on the whole image (both masked \& unmasked areas), it performs a joint prediction and reconstruction task.}
    \label{table:pred_recons}
\vspace{-10pt}
\end{table}

\subsubsection{Prediction Target}\label{sec:pred-target}

Table~\ref{table:pred_type} compares the effects of different prediction targets. Several observations can be drawn as follows:
\begin{itemize}
\vspace{-5pt}
    \item The three losses of $\ell_1$, smooth-$\ell_1$, and $\ell_2$ perform similarly well;
\vspace{-5pt}
    \item Carefully defined classes by color clustering \cite{chen2020imagegpt} or tokenization~\cite{bao2021beit} perform slightly worse than ours; 
\vspace{-5pt}
    \item A simple color discretization approach by channel-wise equal-sized bins (proposed as an alternative option) performs competitive to $\ell_1$ loss, but it requires a careful tuning of the bin number (e.g., 8-bin).
\vspace{-5pt}
\end{itemize}

It reveals that it is not necessary to align the target of masked image modeling to be the same classification based as masked language modeling. It is good to align the approach to the own nature of visual signals.

\paragraph{Prediction or reconstruction?} While both auto-encoders and masked image modeling approaches learn a network by recovering the original signals, they are built on different philosophies of \emph{visible signal reconstruction} and \emph{prediction of invisible signals}. In our framework, we can instantiate a reconstruction task by also regress the raw pixel values of visible patches in the input.

Table~\ref{table:pred_recons} compares the approach which predicts only the masked area as in our default setting and an alternative to recover both masked and unmasked area. The approach predicting the masked area performs significantly better than that recovering all image pixels as 82.8\% vs. 81.7\%. This implies that the two tasks are fundamentally different in their internal mechanisms, and the task to \emph{predict} might be a more promising representation learning approach.

\begin{table}\small
    \centering
    \begin{tabular}{c|c|c}
    \toprule
    Loss & Pred. Resolution & Top-1 acc (\%)\\ 
    \hline
    \hline
    \multicolumn{3}{c}{Classification} \\
    \hline
    8-bin & $192^2$ & 82.7 \\
    8-bin & $48^2$ & 82.7\\
    \hline
    256-bin & $192^2$ & N/A \\
    256-bin & $48^2$ & 82.3\\
    \hline
    iGPT cluster & $192^2$ & N/A \\
    iGPT cluster & $48^2$ & 82.4 \\
    \hline
    BEiT & - & 82.7 \\
    \hline
    \hline
    \multicolumn{3}{c}{Regression} \\
    \hline
    $\ell_2$ & $192^2$ & 82.7\\
    smooth-$\ell_1$ & $192^2$ & 82.7 \\
    $\ell_1$ & $192^2$ & 82.8\\
    $\ell_1$ & $48^2$ & 82.7\\
    $\ell_1$ & $6^2$ & 82.3\\
    \bottomrule
    \end{tabular}
    \vspace{-7pt}
    \caption{Ablation on different prediction targets.}
    \label{table:pred_type}
    \vspace{-10pt}
\end{table}

\begin{table}\small
    \centering
    \addtolength{\tabcolsep}{-5.5pt}
    \begin{tabular}{c|c|ccc}
    \toprule
    \multirow{2}{*}{Methods} & Input & Fine-tuning & Linear eval & Pre-training  \\
    & Size & Top-1 acc (\%) & Top-1 acc (\%) & costs \\
    \hline
    Sup. baseline~\cite{touvron2021training}& $224^2$ & 81.8 & - & - \\
    \hline
    DINO~\cite{caron2021emerging} & $224^2$ & 82.8 & 78.2 & 2.0$\times$ \\
    MoCo v3~\cite{chen2021mocov3} & $224^2$ & 83.2 & 76.7 & 1.8$\times$ \\
    ViT~\cite{dosovitskiy2020vit}& $384^2$ & 79.9 & - & $\sim$4.0$\times$ \\
    BEiT~\cite{bao2021beit} & $224^2$ & 83.2 & 56.7 & 1.5$\times^\dag$\\
    \hline 
    Ours & $224^2$ & \textbf{83.8} & 56.7 & 1.0$\times$ \\
    \bottomrule
    \end{tabular}
    \vspace{-7pt}
    \caption{System-level comparison using ViT-B as the encoder. Training costs are counted in relative to our approach. $^\dag$ BEiT requires an additional stage to pre-train dVAE, which is not counted.}
    \label{table:vit_comparison}
    \vspace{-10pt}
\end{table}

\subsection{Comparison to Previous Approaches on ViT-B}

As previous works~\cite{caron2021emerging,bao2021beit} performed experiments on the ViT architectures, for fair comparison, we also conduct experiments using the ViT-B architecture.

In pre-training, 800 epochs with a cosine learning rate scheduler and a 20-epoch linear warm-up procedure are employed. All other hyper-parameters strictly follow the same settings as in the ablation study, except that we use a $224^2$ input resolution to be the same as in previous approaches.
In fine-tuning, we adopt a layer-wise learning rate decay of 0.65 following \cite{bao2021beit}, and keep all other settings strictly the same as in our ablation study. 
In linear probing, we follow~\cite{bao2021beit} to choose an inter-mediate layer of ViT-B which produces the best linear probing accuracy. 100-epoch training with a 5-epoch linear warm-up step is employed.

Table~\ref{table:vit_comparison} compares our approach to previous ones on both metrics of fine-tuning and linear probing using ViT-B. Our approach achieves a top-1 accuracy of 83.8\% by fine-tuning, which is +0.6\% higher than previous best approach~\cite{bao2021beit}. Also note that our approach reserves the highest training efficiency than others thanks to its simplicity, that it is 2.0$\times$, 1.8$\times$, $\sim$4.0$\times$, and 1.5$\times$ more efficient than that of DINO~\cite{caron2021emerging}, MoCo v3~\cite{chen2021mocov3}, ViT~\cite{dosovitskiy2020vit}, and BEiT~\cite{bao2021beit} (not counting the time for dVAE pre-training), respectively.

Though our main focus is to learn representations that are better for fine-tuning, we also report the linear probing accuracy of different approaches for reference. 

\begin{table}\small
    \centering
    \addtolength{\tabcolsep}{-4.pt}
    \begin{tabular}{c|ccc|ccc}
    \toprule
    Methods & Pre-train & Fine-tune & Backbone  & Top-1 acc (\%) & Param \\
    \hline
    Sup.& $192^2$& $224^2$ & Swin-B  & 83.3 & 88M\\
    Sup.& $192^2$& $224^2$ & Swin-L  & 83.5 & 197M\\
    Sup.& $192^2$& $224^2$ & SwinV2-H  & 83.3 & 658M \\
    \hline 
    Ours & $192^2$& $224^2$ & Swin-B  & 84.0 & 88M\\
    Ours & $192^2$& $224^2$ & Swin-L  & 85.4 & 197M \\
    Ours & $192^2$& $224^2$ & SwinV2-H  & 85.7 & 658M \\
    Ours & $192^2$& $512^2$ & SwinV2-H  & 87.1 & 658M\\
    \hline
    Ours & $192^2$& $640^2$ & SwinV2-G  & 90.2 & 3.0B\\
    \bottomrule
    \end{tabular}
    \vspace{-7pt}
    \caption{Scaling experiments with Swin Transformer as backbone architectures. All our models are pre-trained with input of $192^2$. Different to other models, Swin-G is trained on a privately collected ImageNet-22K-ext dataset, with details described in~\cite{swinv2}.}
    \label{table:swin_comparison}
    \vspace{-10pt}
\end{table}

\subsection{Scaling Experiments with Swin Transformer}

We adopt Swin Transformer of different model sizes for experiments, including Swin-B, Swin-L, SwinV2-H, and SwinV2-G~\cite{swinv2}. To reduce experimental overheads, we adopt a smaller image size of $192^2$ in pre-training, and a step learning rate scheduler that the experiments of different training lengths can reuse model training of the first step. The base learning rate of the first learning rate step is set 4e-4 and lasts for 7/8 of the total training epochs. The learning rate is divided by 10 for the remaining epochs. For model sizes of H and G, we use the variants introduced in~\cite{swinv2}, which have stronger stability than the original version. All models use the ImageNet-1K dataset for training, except that SwinV2-G uses a larger and privately collected ImageNet-22K-ext dataset, as detailed in~\cite{swinv2}.

When using ImageNet-1K for pre-training, all models are trained by 800 epochs, with most other hyper-parameters following that in ablations. In fine-tuning, a larger image size of $224^2$ is employed. For SwinV2-H, we also consider a larger resolution of $512^2$. The training length of fine-tuning is set 100-epoch, except for SwinV2-H where 50-epoch is used. The layer-wise learning rate decay is set as 0.8, 0.75, and 0.7 for Swin-B, Swin-L, and SwinV2-H, respectively. Other fine-tuning hyper-parameters follow that in ablation.

Table~\ref{table:swin_comparison} lists the results of our approach with different model sizes, compared to the supervised counterparts. With \emph{SimMIM} pre-training, all of Swin-B, Swin-L, and SwinV2-H achieve significantly higher accuracy than their supervised counterparts. In addition, the SwinV2-H model with a larger resolution of $512^2$ achieves 87.1\% top-1 accuracy on ImageNet-1K, which is the highest number among methods that use ImageNet-1K data only.

While all previous billion-level vision models rely on Google's JFT-3B dataset for model training~\cite{zhai2021scaling,riquelme2021scaling,dai2021coatnet}, the proposed \emph{SimMIM} approach is used to aid the training of a 3B SwinV2-G model~\cite{swinv2} by using $\sim$40$\times$ smaller data than that of JFT-3B. It achieves strong performance on four representative vision benchmarks: 84.0\% top-1 accuracy on  ImageNet-V2 classification~\cite{recht2019imagenetv2}, 63.1/54.4 box/mask mAP on COCO object detection~\cite{chen2019htc,lin2014coco}, 59.9 mIoU on ADE20K semantic segmentation~\cite{zhou2018ade,xiao2018upernet}, and 86.8\% top-1 acc on Kinetics-400 action recognition~\cite{kay2017kinetics,liu2021video}. More details are described in~\cite{swinv2}.

\begin{figure}
    \centering
    \includegraphics[width=\linewidth]{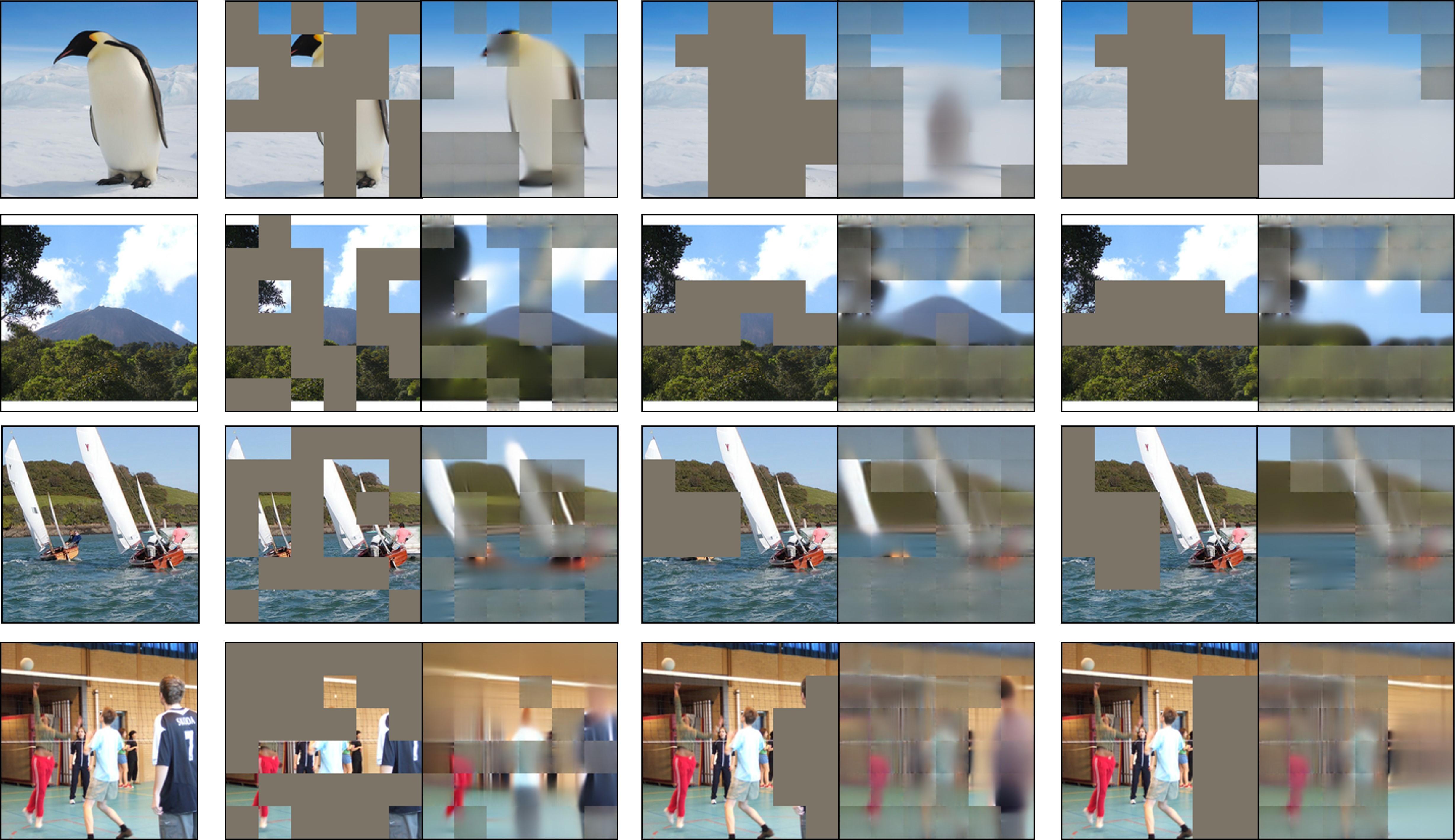}
    \vspace{-20pt}
    \caption{Recovered images using three different mask types (from left to right): random masking, masking most parts of a major object, and masking the full major object.}
    \label{fig:diff_mask}
    \vspace{-10pt}
\end{figure}

\begin{figure}
    \centering
    \includegraphics[width=\linewidth]{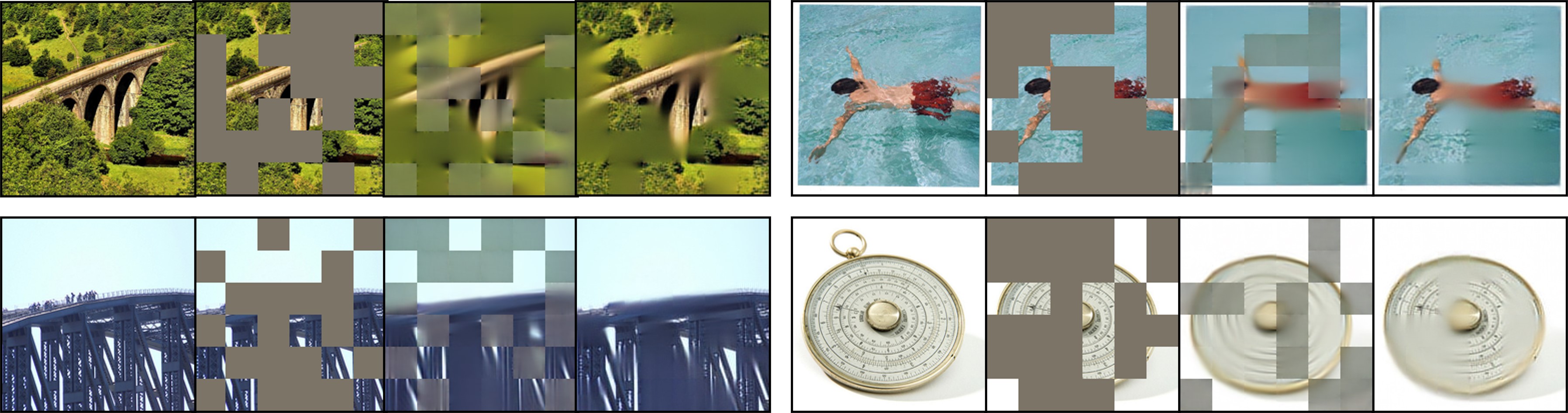}
    \vspace{-20pt}
    \caption{Recovered images by two different losses of predicting only the masked area or reconstructing all image area, respectively. For each batch, images from left to right are raw image, masked image, prediction of \emph{masked patches only}, and reconstruction of \emph{all patches}, respectively.}
    \label{fig:vis_perform_area}
    \vspace{-1em}
\end{figure}

\subsection{Visualization}

In this section, we attempt to understand the proposed approach as well as some critical designs through visualizations. All example images are from the ImageNet-1K validation set.

\paragraph{What capability is learned?} Figure~\ref{fig:diff_mask} shows the recovered images with several human-designed masks, to understand what capability is learnt through masked image modeling. The human-designed masks (from left to right) consist of a random mask, a mask to remove most parts of a major object, and a mask to remove all of the major object, respectively. We can draw the following observations: 1) by random masking moderate parts of the major object, both the shape and texture of masked parts can be well recovered, as shown by the penguin, the mountain, the sailboat, and the persons. On the unmasked area, there is a severe checkerboard artifact due to that the recovery of unmasked area is not learnt during training; 2) by masking most parts of a major object (larger than 90\%), the model can still predict an existence of object by the negligible clues; 3) when the objects are fully masked out, the masked area will be inpainted with background textures.

These observations indicate that the approach has learnt strong reasoning ability of objects, and the ability is not due to memorization of image identities or the simple copying of nearby pixels.

\vspace{-.5em}
\paragraph{Prediction v.s. reconstruction} 
We have shown the comparison of the representations learnt by a masked prediction task (our approach), and a joint masked prediction and visible signal reconstruction task in Table~\ref{table:pred_recons}, which reveals that the pure masked prediction task performs significantly better. Figure~\ref{fig:vis_perform_area} compares the recovery effects by two approaches. It shows that the latter approach makes better looking, however, probably the model capacity is wasted at the recovery of the unmasked area which may not be that useful for fine-tuning.

\begin{figure}
    \centering
    \includegraphics[width=\linewidth]{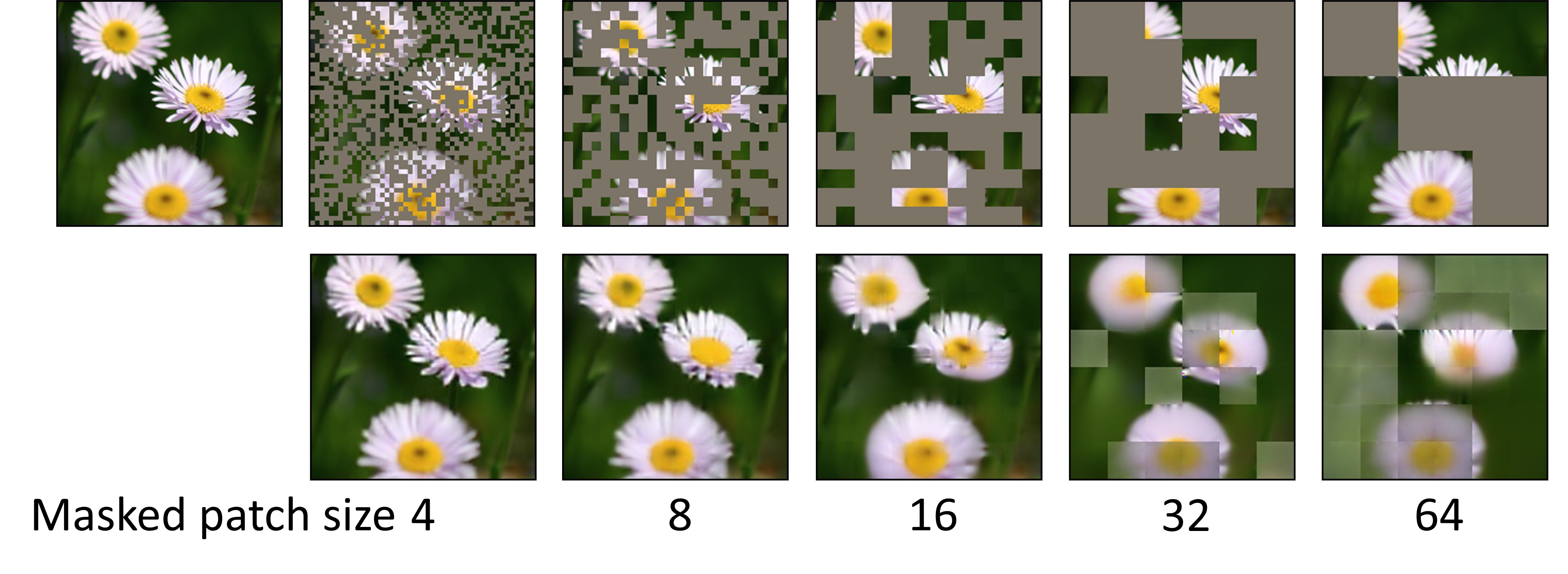}
    \vspace{-25pt}
    \caption{An example of recovered image using masked patch sizes of 4, 8, 16, 32 and 64, and a fixed masked ratio of 0.6.}
    \label{fig:vis_masked_patch}
    \vspace{-1em}
\end{figure}

\vspace{-.5em}

\paragraph{Effects of masked patch size} 

Figure~\ref{fig:vis_masked_patch} shows the recovery of an image with different masked patch size under a fixed masking ratio of 0.6. It can be seen that the details can be much better recovered when the masked patch size is smaller, however, the learnt representations transfer worse. Probably, with smaller patch size, the prediction task can be easily accomplished by close-by pixels or textures.

\section{Conclusion}
\label{sec:conclusion}

This paper presents a simple yet effective self-supervised learning framework, SimMIM, to leverage masked image modeling for representation learning. This framework is made as simple as possible: 1) a random masking strategy with a moderately large masked patch size; 2) predicting raw pixels of RGB values by direct regression task; 3) the prediction head can be as light as a linear layer. We hope our strong results as well as the simple framework can facilitate future study of this line, and encourage in-depth interaction between AI fields.

\section*{Acknowledgement}

We thank many colleagues at Microsoft for their help, in particular, Li Dong, Furu Wei, Eric Chang, Lidong Zhou, Jing Tao, Aaron Zhang, Edward Cui, Peng Cheng and Fan Yang for useful discussion and the help on GPU resources and datasets.


\section*{\Large{Appendix}}

\appendix

\begin{table}[h]\small
    \centering
    \begin{tabular}{c|c|c|c}
    \toprule
    Model & Swin-B & Swin-L & SwinV2-H \\
    \hline
    Base Channel & 128 & 192 & 352 \\
    Depths & \{2, 2, 18, 2\} & \{2, 2, 18, 2\} & \{2, 2, 18, 2\}\\
    Params & 88M & 197M & 658M \\
    \hline
    \multicolumn{4}{c}{\emph{Pre-training}}\\
    \hline
    Input Size & 192 & 192 & 192 \\
    Window Size & 6 & 12 & 12 \\
    FLOPs & 11.3G & 26.0G & 86.2G \\
    \hline
    \multicolumn{4}{c}{\emph{Fine-tuning}} \\
    \hline
    Input Size & 224 & 224 & 224 \\
    Window Size & 7 & 14 & 14 \\
    FLOPs & 15.4G & 35.8G & 118.1G \\
    \bottomrule
    \end{tabular}
    \caption{Detailed architecture specifications.}
    \label{table:Swin_variants}
    \vspace{-10pt}
\end{table}

\section{Detailed Architectures}
The detailed architecture specifications are shown in Table~\ref{table:Swin_variants}, where an input image size of $192\times192$ is used for pre-training and $224\times224$ is used in fine-tuning.

\section{The Effect of Learning Rate Schedulers}

In our ablation study, we follow common practice~\cite{dosovitskiy2020vit,bao2021beit} to use a \emph{cosine} learning rate scheduler. In our scaling up experiments, we adopt a \emph{step} learning rate scheduler to reduce experimental overheads of potentially studying the effects of different training lengths.

In this section, we investigate the effects of different schedulers on fine-tuning accuracy. Both schedulers adopt 10-epoch linear warm-up. For the \emph{step} learning rate scheduler, the base learning rate is set as 8e-4, and is decayed by a factor of 10 at 90\% and 95\% of the total training length. For this comparison, we follow the default settings used in ablation, except that the scheduler is changed. As shown in Table~\ref{table:lr_schedule}, the \emph{step} scheduler performs marginally better than the \emph{cosine} scheduler, by +0.1\% using a 100-epoch pre-training, and by +0.3\% using a longer 300-epoch training procedure.

\begin{table}\small
    \centering
    \begin{tabular}{c|c|c}
    \toprule
    lr scheduler & 100 epochs & 300 epochs\\
    \hline
    cosine & 82.8 & 83.0\\
    step & 82.9 & 83.3\\
    \bottomrule
    \end{tabular}
    \caption{The effects of different learning rate schedulers.}
    \label{table:lr_schedule}
\end{table}

\section{Results on Downstream Tasks}
In this section, we add more results on several down-stream tasks, including iNaturalist (iNat) 2018 classification, COCO object detection and ADE20K semantic segmentation.

\subsection{Detailed Settings}

\paragraph{iNaturalist 2018 classification}
iNaturalist~\cite{van2018inaturalist} 2018 is a long-tail image classification dataset with more than 8,000 categories. It includes 437,513 training images and 24,426 validation images. We fine-tune the pre-trained models using an AdamW optimizer by 100 epochs. The fine-tuning hyper-parameters are: a batch size of 2048, a base learning rate of 1.6e-2, a weight decay of 0.05, $\beta_{1}=0.9$, $\beta_{2}=0.999$, a stochastic depth~\cite{huang2016deep} ratio of 0.1, and a layer-wise learning rate decay of 0.9. We follow the same data augmentation strategies used in \cite{bao2021beit}, including RandAug~\cite{cubuk2020randaugment}, Mixup~\cite{zhang2017mixup}, Cutmix~\cite{yun2019cutmix}, label smoothing~\cite{szegedy2016rethinking}, and random erasing~\cite{zhong2020random}.

\paragraph{COCO object detection}
A Mask-RCNN\cite{Mask-rcnn} framework is adopted and all models are trained with a 3$\times$  schedule (36 epochs). We utilize an AdamW\cite{kingma2014adam} optimizer with a learning rate of 6e-5, a weight decay of 0.05, and a batch size of 32. Following \cite{simple_copy_paste}, we employ a large jittering augmentation (1024 $\times$ 1024 resolution, scale range [0.1, 2.0]). The window size for Swin-B is set to 7 and that for Swin-L and SwinV2-H models is 14.

\paragraph{ADE20K semantic segmentation}
Following \cite{liu2021swin}, An UPerNet framework\cite{xiao2018upernet} is used following \cite{liu2021swin}. We use an AdamW\cite{kingma2014adam} optimizer using the following hyper-parameters: a weight decay of 0.05, a batch size of 32, a layer-wise decay rate of 0.9, and a learning rate searching from 1e-4 and 3e-4. All models are trained for 80K iterations with an input resolution of 512$\times$512 and a window size of 20. In inference, a multi-scale test using resolutions that are [0.75, 0.875, 1.0, 1.125, 1.25]$\times$ of 512$\times$2048 is employed.

For ADE20K experiments, we initialized the segmentation models using model weights after supervised fine-tuning on ImageNet-1K, because its performance is superior to using the self-supervised pre-trained weights directly.

\begin{table}
    \centering
    \addtolength{\tabcolsep}{-2.5pt}
    \begin{tabular}{c|c|c|c|c|c}
    \toprule
    \multicolumn{2}{c|}{\multirow{2}{*}{Head}} & ImageNet & iNat-2018 & COCO & ADE20K\\
    \multicolumn{2}{c|}{} & Top-1 Acc & Top-1 Acc & mAP$^{\text{box}}$ & mIoU \\
    \hline
    \multicolumn{2}{c|}{Linear} & 82.8 & 75.2 & 49.9 & 50.0 \\
    \multicolumn{2}{c|}{2-layer MLP} & 82.8 & 75.0 & 50.1 & 49.9 \\
    \multicolumn{2}{c|}{inverse Swin-T} & 82.4 & 74.9 & 49.8 & 49.4 \\
    \multicolumn{2}{c|}{inverse Swin-B} & 82.5 & 75.0 & 49.8 & 49.0 \\
    \bottomrule
    \end{tabular}
    \caption{More ablation studies on prediction head designing using iNat-2018, COCO and ADE20K.}
    \label{table:rebuttal_ablation_pred_head}
\end{table}

\begin{table}\small
    \centering
    \addtolength{\tabcolsep}{-2.5pt}
    \begin{tabular}{c|c|c|c|c|c}
    \toprule
    \multirow{2}{*}{Loss} & Pred. & ImageNet & iNat-2018 & COCO & ADE20K \\
    & Resol. & Top-1 Acc & Top-1 Acc & mAP$^{\text{box}}$ & mIoU \\
    \hline
    8-bin & $48^2$ & 82.7 & 75.3 & 50.0 & 49.7 \\
    256-bin & $48^2$ & 82.3 & 74.6 & 49.7 & 49.3 \\
    iGPT & $48^2$ & 82.4 & 75.0 & 49.6 & 49.1 \\
    BEiT & - & 82.7 &  75.2 & 50.1 & 48.8 \\
    $\ell_1$ & $192^2$ & 82.8 & 75.2 & 49.9 & 50.0 \\
    \bottomrule
    \end{tabular}
    \caption{More ablation studies on prediction targets using iNat-2018, COCO and ADE20K.}
    \label{table:rebuttal_ablation_pred_target}
\end{table}

\begin{figure}
    \centering
    \includegraphics[width=1.0\linewidth]{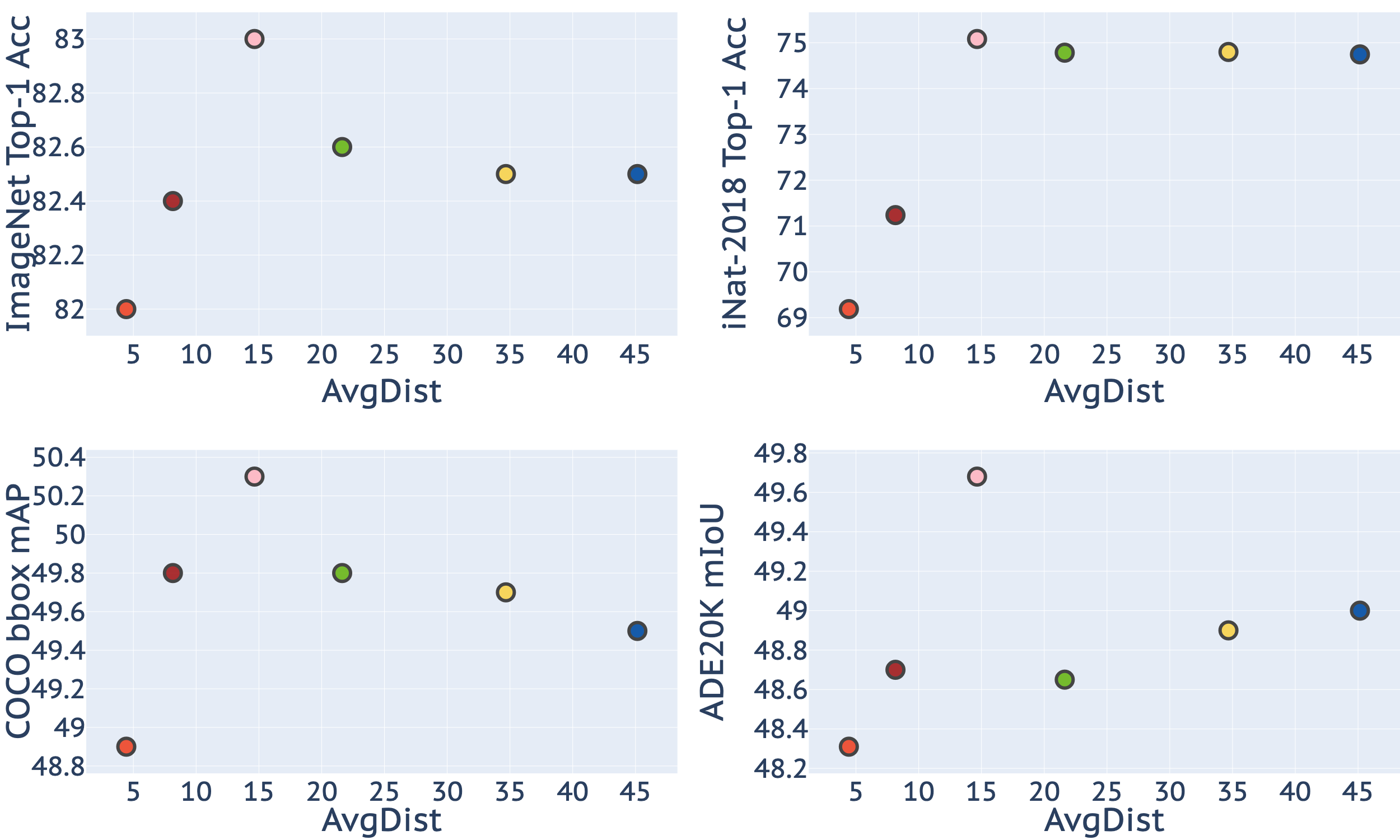}
    \caption{\emph{AvgDist} (averaged distance of masked pixels to the nearest visible pixels) w.r.t. performance on ImageNet, iNat-2018, COCO and ADE20K.}
    \label{fig:avgdist_results}
\end{figure}

\subsection{Ablation Studies}
Table~\ref{table:rebuttal_ablation_pred_head} and ~\ref{table:rebuttal_ablation_pred_target} ablates the designs in SimMIM on the above additional down-stream tasks. We also copy the results of ImageNet-1K from the main body to these tables for reference. 

Table~\ref{table:rebuttal_ablation_pred_head} indicates that a lighter head (linear, 2-layer) is consistently better than the heavier heads (e.g. inverse Swin-T) on most tasks: +0.4\% on ImageNet-1K, +0.3\% on iNat-2018, and +0.6 on ADE20K. Table~\ref{table:rebuttal_ablation_pred_target} suggests that our presented regression based prediction target ($\ell_1$) could achieve on par or better performance than the well designed classification based ones. 

We also use these additional down-stream tasks to verify different masking strategies, as shown in Figure~\ref{fig:avgdist_results}. It turns out that the observations in Figure~3 of the main paper also hold: 1) the \emph{AvgDist} measure is a good indicator for the learning effectiveness of masked image modeling; 2) an \emph{AvgDist} of $15$ is empirically good for masked image modeling.

\begin{table}\small
    \centering
    \begin{tabular}{c|c|c|c|c}
    \toprule
    \multirow{3}{*}{Backbone} & \multicolumn{2}{|c|}{Sup.} & \multicolumn{2}{|c}{Ours} \\
    \cline{2-5}
    & COCO & ADE20K & COCO & ADE20K \\
    & mAP$^{\text{box}}$ & mIoU & mAP$^{\text{box}}$ & mIoU \\
    \hline
    Swin-B & 50.2 & 50.4 & 52.3 & 52.8 \\
    Swin-L & 50.9 & 50.0 & 53.8 & 53.5 \\
    SwinV2-H & 50.2 & 49.8 & 54.4 & 54.2 \\
    \bottomrule
    \end{tabular}
    \caption{Scaling experiments with Swin on COCO and ADE20K.}
    \label{table:scaling_coco_ade}
\end{table}

\subsection{Scaling Experiments}
Table~\ref{table:scaling_coco_ade} shows the scaling performance using COCO object detection and ADE20K semantic segmentation.
On Swin-B, Swin-L, and SwinV2-H, SimMIM achieves +2.1 / +2.9 / +4.2 mAP$^{\text{box}}$ and +2.4 / +3.5 / +4.4 mIoU higher accuracy than its supervised counterparts, respectively. It indicates the broad effectiveness of the SimMIM approach. It also suggests that larger models benefit more from this approach.

\section{More Results on Channel-wise Bin Color Discretization}

Table~\ref{table:supp_pred_type} shows more results of using \emph{channel-wise bin color discretization} as the prediction target, by varying bin numbers and prediction resolutions. We notice that the best accuracy for different bin numbers are achieved at different prediction resolutions: the 2-bin and 4-bin targets reach the best accuracy at a resolution of $192^2$, and all other bin numbers reach the best accuracy at a low prediction resolution of $6^2$. These results imply a moderately fine-grained target is encouraged for this classification based approach. 

\begin{table}\small
    \centering
    \addtolength{\tabcolsep}{-1.pt}
    \begin{tabular}{c|c|c|c|c|c|c}
    \toprule
    \multirow{2}{*}{Pred. Resolution} & \multicolumn{6}{c}{Bin Num. (Top-1 acc \%)} \\ 
    \cline{2-7}
    & 2 & 4 & 8 & 16 & 32 & 256 \\
    \hline
    $6^2$ & 82.5 & 82.7 & \textbf{82.8} & \textbf{82.9} & \textbf{82.8} & \textbf{82.4}\\
    $48^2$ & 82.5 & 82.8 & 82.7 & 82.6 & 82.5 & 82.3\\
    $192^2$ & \textbf{82.7} & \textbf{82.9} & 82.7 & 82.7 & N/A & N/A\\
    \bottomrule
    \end{tabular}
    \caption{More results of using \emph{channel-wise bin color discretization} as the prediction target, by varying bin numbers and prediction resolutions. Swin-B and 100-epoch pre-training are used.}
    \label{table:supp_pred_type}
\end{table}

\section{SimMIM with ConvNets}
With the remarkable performance of SimMIM on Vision Transformers, we want to verify its effectiveness on versatile architectures. Here we adopt ResNet-50$\times$4 as the base architecture. The overall training setup remains the same as that of Swin-Base. We use masked tokens to replace the original features after the stem of a $3\times3$ convolution of $stride=2$ followed by a 2$\times$2 max-pooling operator.

On ResNet-50$\times$4, SimMIM achieves 81.6\% top-1 accuracy on ImageNet-1K validation set using 300-epoch pre-training and 100-epoch fine-tuning, outperforming the supervised counterpart by +0.9\% (vs. 80.7\%). This indicates the generality of SimMIM.

{\small
\bibliographystyle{ieee_fullname}
\bibliography{egbib}
}

\end{document}